\documentclass[lettersize,journal]{IEEEtran}
\usepackage{amsmath,amsfonts}
\usepackage{algorithmic}
\usepackage{algorithm}
\usepackage{array}
\usepackage[caption=false,font=normalsize,labelfont=sf,textfont=sf]{subfig}
\usepackage{textcomp}
\usepackage{stfloats}
\usepackage{url}
\usepackage{verbatim}
\usepackage{graphicx}
\usepackage{cite}
\usepackage[table,dvipsnames]{xcolor}
\usepackage{colortbl}
\definecolor{RoyalBlue}{RGB}{34,139,34}
\definecolor{LightBlue}{RGB}{173,216,230}
\definecolor{mypink}{RGB}{255,105,180}
\usepackage[colorlinks=true, linkcolor=mypink, citecolor=mypink, urlcolor=mypink]{hyperref}
\makeatletter
\def\@cite#1#2{{\hypersetup{citecolor=blue}\textcolor{blue}{[{#1\if@tempswa , #2\fi}]}}}
\makeatother
\usepackage{booktabs}
\usepackage{multirow}
\begin{document}

\title{DCCS-Det: Directional Context and Cross-Scale-Aware Detector for Infrared Small Target}

\author{Shuying~Li, Qiang~Ma, San~Zhang, Chuang~Yang,~\IEEEmembership{Member,~IEEE}
	\thanks{Shuying Li and Qiang Ma are with the School of Automation, Xi’an University of Posts and Telecommunications, Xi’an 710121, China, and also with Shanghai Artificial Intelligence Laboratory, Shanghai 200232, China (e-mail: lishuying@xupt.edu.cn; maqiang123@stu.xupt.edu.cn).}%
	\thanks{San Zhang is with the Key Laboratory of Cyberspace Security, Ministry of Education of China and Henan Key Laboratory of Cyberspace Situation Awarenes, Zhengzhou 450001, China (e-mail: zhangsan@xupt.edu.cn).}%
	\thanks{Chuang Yang is with the Department of Electrical and Electronic Engineering, The Hong Kong Polytechnic University, Hong Kong SAR (e-mail:omtcyang@gmail.com).}
	\thanks{Corresponding author: Chuang Yang.}
}

\markboth{~}%
{Shell \MakeLowercase{\textit{et al.}}: A Sample Article Using IEEEtran.cls for IEEE Journals}


\maketitle

\begin{abstract}
Infrared small target detection (IRSTD) is critical for applications like remote sensing and surveillance, which aims to identify small, low-contrast targets against complex backgrounds. However, existing methods often struggle with inadequate joint modeling of local-global features (harming target-background discrimination) or feature redundancy and semantic dilution (degrading target representation quality). To tackle these issues, we propose DCCS-Det (Directional Context and Cross-Scale Aware Detector for Infrared Small Target), a novel detector that incorporates a Dual-stream Saliency Enhancement (DSE) block and a Latent-aware Semantic Extraction and Aggregation (LaSEA) module. The DSE block integrates localized perception with direction-aware context aggregation to help capture long-range spatial dependencies and local details. On this basis, the LaSEA module mitigates feature degradation via cross-scale feature extraction and random pooling sampling strategies, enhancing discriminative features and suppressing noise. Extensive experiments show that DCCS-Det achieves state-of-the-art detection accuracy with competitive efficiency across multiple datasets. Ablation studies further validate the contributions of DSE and LaSEA in improving target perception and feature representation under complex scenarios.  \href{https://huggingface.co/InPeerReview/InfraredSmallTargetDetection-IRSTD.DCCS}{DCCS-Det Official Code is Available Here!}
\end{abstract}

\begin{IEEEkeywords}
 Infrared small target detection(IRSTD), target-background discrimination, feature redundancy, long-range spatial dependencies, cross-scale feature extraction.
\end{IEEEkeywords}

\section{INTRODUCTION}
\IEEEPARstart{I}{nfrared} small target detection (IRSTD) is a key technique in remote sensing surveillance and military early-warning applications. It aims to  detect and locate small and dim targets against complex background. Due to the limitations of long-range imaging, these targets always occupy only a few pixels.
Meanwhile, atmospheric scattering, optical blur, and sensor noise further reduce the signal-to-noise ratio (SNR) and the contrast between the target and the background. As a result, the image lacks detailed texture and structural information. Furthermore, background clutter from clouds, man-made structures, and vegetation tends to obscure and submerge weak targets~\cite{9714770,kou2023infrared}. These factors make reliable detection particularly challenging in practical IRSTD scenarios. 

In response to these challenges, research on IRSTD advanced a variety of classical methodologies, predominantly grounded in hand-crafted features and explicit prior assumptions. Broadly, these approaches can be categorized into three principal classes. The first comprises image filtering and background suppression methods~\cite{deshpande1999max,tom1993morphology,zhao2014bilateral}. In these methods, infrared imagery is processed through filtering or background estimation to accentuate the minute target signatures while attenuating the relatively homogeneous background. The second class draws inspiration from the human visual system methods~\cite{6479296,7460907,wei2016multiscale}, leveraging localized contrast enhancement to exploit the perceptual sensitivity to salient deviations between a target and its surrounding context. The third category encompasses background modeling based on low-rank and sparse decomposition methods~\cite{6595533,7999276,9203993}. These methods decompose a raw infrared frame into a low-rank background component and a sparse target component, thereby separating the target from the background. While these approaches demonstrate satisfactory performance in straightforward environments, they suffer from reduced reliability when dealing with complex backgrounds due to restrictive prior constraints.

In contrast to traditional approaches, deep learning-based methods were proposed to address these problems, many of which leverage Convolutional Neural Networks (CNNs)~\cite{9423171,9864119,9314219,9989433,10011452,zhang2022exploring,zhang2024irprunedet}. For example, Dai et al. constructed ACM-Net~\cite{9423171} and ALC-Net~\cite{9314219} based on the U-shaped architecture~\cite{ronneberger2015u} and designed a multi-level fusion method for asymmetric contextual modulation features. Li et al. designed DNA-Net~\cite{9864119} based on U-Net++~\cite{zhou2018unet++} to progressively integrate high-level and low-level features. Alongside this, FC3-Net~\cite{zhang2022exploring} employed fine-detail guided multi-level feature aggregation to counteract detail degradation from successive downsampling. Concurrently, IRPruneDet~\cite{zhang2024irprunedet} achieved notable parameter reduction while preserving detection capability through wavelet-domain sparse constraints. However, although these CNNs achieve promising detection performance, they often overlook the importance of global contextual information for precise target discrimination. It leads to high consistency between small targets and interfering pixels in infrared images, which limits improvement in overall detection performance.

To effectively capture the semantic correlation between targets and complex backgrounds, many works began to explore hybrid architectures that incorporate Transformers~\cite{9912644,10011449,10024907,10219645,10486932,zhang2024irsam,zhang2025saist}. They utilize self-attention mechanisms to model global feature correlations, compensating for the limited receptive field of CNNs. FTC-Net~\cite{9912644} drew on the design philosophy of Swin Transformer~\cite{9710580} and combined the local features of CNNs with the global context of Swin Transformer for feature modeling. However, the network performed feature fusion only during the output phase, which hinders the establishment of an effective guiding mechanism between the target and background semantics. Additionally, MTU-Net~\cite{10011449} and SCTransNet~\cite{10486932} used Transformers to enhance global dependency modeling across different feature levels or scales, which in turn improved the performance of feature fusion. More recently, IRSAM~\cite{zhang2024irsam} incorporated Perona-Malik diffusion filtering alongside granularity-aware decoding, jointly achieving structural preservation and noise attenuation. SAIST~\cite{zhang2025saist} harnessed cross-modal contrastive learning to couple textual cues with visual representations, improving detection reliability amid intricate backgrounds. However, these frameworks often struggle to represent the latent target semantic regions in the deeper layers of the network. Consequently, the fusion phase is unable to leverage the latent semantics to exert a guiding and suppressive effect on the background semantics. Moreover, excessive computational complexity further impacted their performance.

To address these challenges, we propose DCCS-Det. This detector employs the DSE block to construct an auxiliary encoder, which provides information supplementation to the primary encoder, thereby enabling joint feature modeling between them. Furthermore, the LaSEA module enhances deep-layer target semantics and provides semantic guidance to the decoder fusion. Specifically, our DCCS-Det employs two parallel branches, where its primary branch contains multi-level residual blocks (Res-Block), emphasizing local detail extraction in small targets. In parallel, the auxiliary branch includes multi-level DSE Block. It supplements the ability of long-range contextual dependency modeling with linear computational complexity. It also enriches local detail information. Through the collaborative work of the two branches, the boundary difference between the target and interference becomes more significant. Furthermore, considering the lower target resolution and the presence of information redundancy in the deeper layers would make it difficult to provide target semantic guidance for the decoding fusion stage, we introduce the LaSEA module to integrate cross-scale feature fusion with random pooling sampling strategies. It helps capture finer features, enhance the semantic region representations of deep small targets, and suppresses noise. The main contributions of this work are as follows:

1) A DSE block is developed to model the long-range contextual dependency between the target and its surrounding environment, along with the local details of both, reinforcing the boundary difference between the target and interference.

2) A LaSEA module is designed to use cross-scale feature extraction and random pooling sampling strategies, which enhances the representation of target semantic regions and suppresses complex background noise.

3) Based on these two modules, DCCS-Det is constructed to improve target-interference boundary discrimination and enhance deep target semantic expression, which achieves superior performance on multiple public benchmarks.
\section{RELATED WORK}
\subsection{Deep Learning Models for IRSTD}
Compared to traditional IRSTD methods based on manually designed features and prior assumptions, deep learning-based methods can learn and capture complex nonlinear features that are difficult to describe with traditional methods. They also avoid the need to predefine complex feature extraction rules. For instance, researchers explored densely nested interaction strategies~\cite{9989433,10149373} to achieve multi-level interactive representation learning of targets. Specifically, UIU-Net~\cite{9989433} embedded a smaller U-Net~\cite{ronneberger2015u} into a larger U-Net backbone, and augmented the network's feature extraction capability. DMFNet~\cite{10466752} and DC-Net~\cite{10706880} adopted cross-layer feature fusion strategies to achieve accurate processing of targets. In addition to the aforementioned strategies, some studies also explored modeling approaches for preserving target shapes and edges~\cite{zhang2022isnet,10830282,10955237,11106397,yang2023instance}. MMLNet~\cite{10830282} employed a multi-branch mutual-guiding framework that adaptively fused edge, localization, and detection features through edge extraction, target positioning, and lossless encoding.

In recent years, Transformer-based hybrid methods were proposed to establish global perceptual dependencies between targets and backgrounds~\cite{chen2022irstformer,9745054,10559654,11159514}. IRSTFormer~\cite{chen2022irstformer} explicitly modeled correlations among various image regions using multi-level overlapping patch-based transformer architectures. In the coarse-to-fine detection approach, the introduction of the Transformer encoder enabled IAANet~\cite{9745054} to model the internal relationships between pixels in the coarse target region. Some authors also explored transformer-based asymmetric encoder–decoder frameworks. Yang et al. proposed PBT~\cite{10559654}, where the encoding stage focused separately on candidate target responses and background context, and the decoding stage aligned them. In addition, HSTNet~\cite{11159514} performed sparse modeling of the relationship between the target and the background through a spatial-channel sparse transformer.

Recently, researchers investigated vision models with linear complexity~\cite{liu2024vmamba,huang2024localmamba,zhang2025mamba,yue2024medmamba}. A key approach is to employ two-dimensional selective scanning (SS2D), first proposed in VMamba~\cite{liu2024vmamba}, which unfolds image data into one-dimensional sequences along multiple directions before feeding them into a state space model (SSM~\cite{gu2023mamba}). By integrating SS2D, the model can effectively capture long-range dependencies across various tasks such as video understanding, medical image segmentation, and cross-modal vision-language alignment. Examples include LocalMamba~\cite{huang2024localmamba} and MFuser~\cite{zhang2025mamba}. However, when SS2D is directly applied to IRSTD tasks, its performance is limited. Specifically, direct application of generic vision Mamba models is suboptimal for IRSTD, as their downsampling strategies and feature parameterization are not tailored for detecting targets that are only a few pixels in size and lack rich texture. This is mainly due to the small size, low contrast, and ambiguous background commonly found in infrared images. Therefore, unlike the aforementioned methods, we design the DSE block as a key component of the auxiliary branch. It works collaboratively with the primary branch, strengthening the model's capacity for capturing both local details and global context.

\begin{figure*}[!t]
	\centering
	\includegraphics[width=\textwidth]{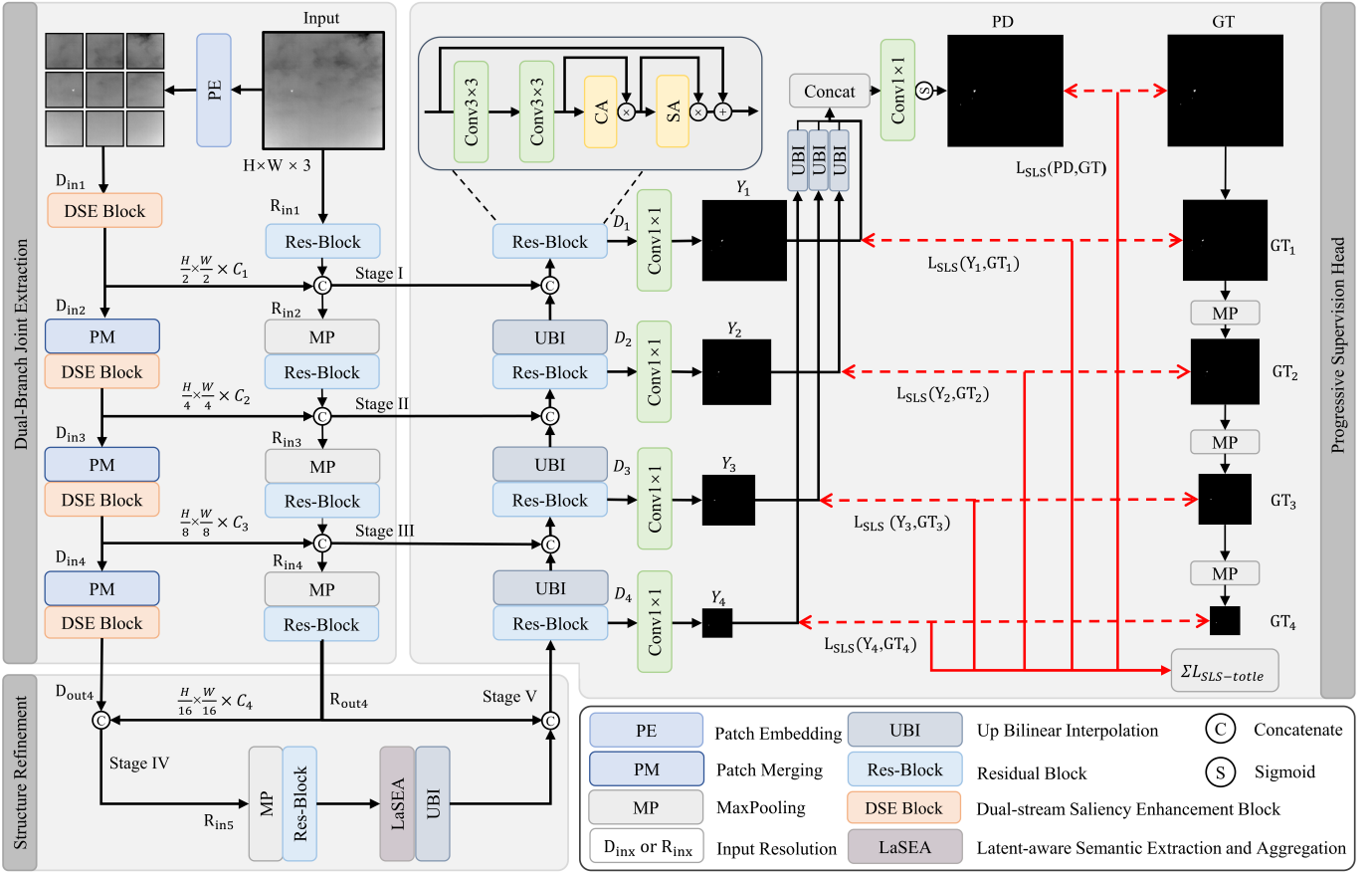}
	\caption{Complete framework of the DCCS-Det method. First, the infrared image is fed into the Dual-Branch Joint Extraction to model features. Then, the Structure Refinement enhances the representation of the target semantics in the deeper layers of the network to guide fusion in the decoder. Finally, the Progressive Supervision Head utilizes the fused features to predict the final result. In the figure, PD represents the predicted result, and GT represents the Ground Truth (actual label).}
	\label{Fig.1}
\end{figure*}

\subsection{Enhanced Fusion Methods in IRSTD}
Feature fusion serves as the bridge connecting encoder and decoder, and its effectiveness directly determines the accuracy of target detection. Therefore, exploring effective fusion enhancement methods for reducing background noise while improving target saliency is of great significance. For example, SeRankDet~\cite{10677425} and EAAU-Net~\cite{tong2021eaau} performed feature integration through cross-layer feature fusion modules to enhance the discriminability of target features. Similarly, SCAFNet~\cite{10745528} strengthened the fusion of different levels of features through a semantic-guided cascade design. Furthermore, CODNet~\cite{11122475} enriched multi-level feature representations through spatial–temporal–frequency domain fusion, improving critical information utilization in the decoding process. To further improve fusion performance, SDS-Net~\cite{11078296} modeled shallow features, deep features, and cross-layer feature correlations separately, significantly enhancing the fused target representation. There are also some works, MAPFF~\cite{10497612} provided convenience for subsequent processing by fully utilizing single-layer resolution-level features and filtering effective information. FDPF-Net~\cite{10994565} aggregated shallow-layer details and deep-layer semantics through feature pyramid fusion module, further enhancing target boundary information. Unlike previous methods, we explore the potential semantic features at deeper network levels. To be specific, after the joint extraction, we further design the LaSEA module in deeper network layers, which efficiently enhances target semantic encoding and suppresses complex background noise. Thus, significant target semantic guidance is provided, making the target features more robust during feature fusion.

\section{METHODOLOGY}
This section details the proposed DCCS-Det for IRSTD, including the complete framework architecture, the DSE Block, and the LaSEA module.

\subsection{Overall Pipeline}
The complete framework of the DCCS-Det model is illustrated in \text{Fig.~\ref{Fig.1}}. It consists of the Dual-Branch Joint Extraction, the Structure Refinement, and the Progressive Supervision Head. Specifically, in \textbf{Dual-Branch Joint Extraction}, an infrared image $\mathbf{I}_1 \in \mathbb{R}^{H \times W \times 3}$ serves as input. In the primary branch, inputs are initially projected into a high-dimensional feature space, resulting in the feature representation $\mathbf{I}_1^{\prime} \in \mathbb{R}^{H \times W \times C^{\prime}}$. The local detail features are subsequently extracted through the concatenation of the Residual Block ($\text{Res-Block}$) and MaxPooling ($\text{MP}$) layers, forming the feature flow $\mathbf{R}_{\text{in}i}$. For the auxiliary branch, the input is first passed through the Patch Embedding (PE) layer, which transforms the image into a high-dimensional feature $\mathbf{I}_1^{\prime\prime} \in \mathbb{R}^{H \times W \times C^{\prime\prime}}$, where $H$ and $W$ represent spatial dimensions (height and width), and $C^{\prime}, C^{\prime\prime}$  indicate channel counts. After passing through the DSE block, the feature flow $\mathbf{D}_{\text{in}i}$ is formed. The primary and auxiliary branches are concatenated at each layer, with the output serving as input for the next layer. The equation is defined as follows:
\begin{equation}
	\mathbf{R}_{\text{in}(i+1)} = \text{Concat}(\mathbf{R}_{\text{in}i}, \mathbf{D}_{\text{in}i}), \quad i = 1, 2, 3, 4,
\end{equation}
where $i$ corresponds to different stages of the model.

In \textbf{Structure Refinement}, the deep features $\mathbf{R}_{\text{in}5}$ are further fed into the $\text{LaSEA}$ module to enhance the structural representation of deep small targets. The equation is defined as follows:
\begin{equation}
	\label{deqn_ex1}
	\mathbf{B} = \text{UBI}(\text{LaSEA} (\text{Res-Block}(\text{MP}(\mathbf{R}_{\text{in}5})))),
\end{equation}
where $\text{UBI}(\cdot)$ denotes upsampling by bilinear interpolation and the feature $\mathbf{B}$ represents the enhanced deep feature output of the Structure Refinement.

In \textbf{Progressive Supervision Head}, during the decoding stage, the features of each layer are first upsampled through UBI, then fused with the feature map at the corresponding resolution. The spatial resolution is progressively restored layer by layer, followed by prediction to generate the confidence map. Here, the decoding features are denoted as $\mathbf{D}_i$, where $i \in \{1, 2, 3, 4\}$ corresponds to different decoding levels, with the number of channels being $\{C, 2C, 4C, 8C\}$, where $C$ is set to 16.
\subsection{Dual-stream Saliency Enhancement Block}
\begin{figure}[!t]
	\centering
	\includegraphics[width=0.48\textwidth]{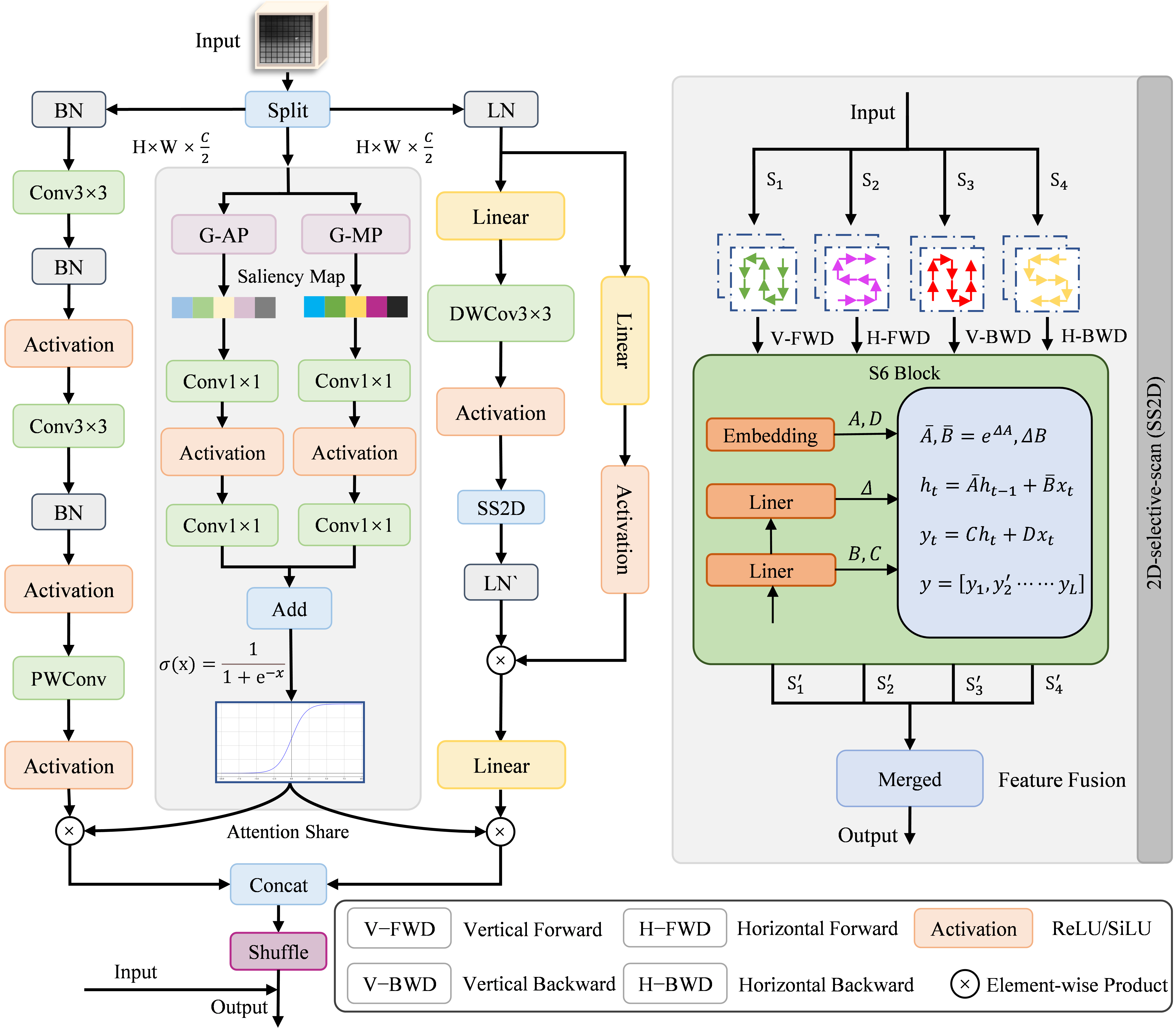}
	\caption{This figure illustrates the architecture of the DSE Block. This module consists of two branches: the left branch enhances local detail features; the right branch employs 2D-Selective-Scan (SS2D) to capture global contextual information with linear complexity. These two feature streams join with the primary branch to achieve collaborative modeling of local details and global dependencies of the target.}
	\label{Fig.2}
\end{figure}
In IRSTD tasks, targets are typically characterized by small spatial dimensions, low contrast, and complex backgrounds, which result in weak saliency and susceptibility to being obscured by background noise. Consequently, it is necessary to perform effective local and global modeling concurrently. In terms of feature extraction, CNNs are constrained by their limited receptive fields, rendering them inadequate for modeling long-range spatial dependencies. Transformer-based hybrid architectures possess global modeling capabilities; however, their computational complexity scales quadratically with the input size. To address this issue, we developed a DSE block that adapts the SS2D~\cite{liu2024vmamba} to IRSTD, providing directional context to supplement the primary branch. As shown in \text{Fig. \ref{Fig.2}}. Specifically, for a feature map $\mathbf{X} \in \mathbb{R}^{H \times W \times C}$, we separate $\mathbf{X}$ in the channel direction forming dual branches. This channel-wise splitting design ensures that the left branch with stacked convolutions captures fine-grained local information crucial for small targets, while the right branch employs SS2D to capture long-range spatial dependencies through four directional scanning paths, encoding directional information into the global context for target-background discrimination. This directional global context, together with the local features from the left branch, forms a comprehensive directional context that supplements the primary branch. This parallel design ensures that both local details and direction-aware global context are preserved without mutual interference. Meanwhile, processing half of the channels with SS2D significantly reduces computational overhead compared to applying SS2D to all channels, achieving a better accuracy-efficiency trade-off. The equation is defined as follows:
\begin{equation}
	(\mathbf{X}^{(L)}, \mathbf{X}^{(R)}) = \text{Split}(\mathbf{X}).
\end{equation}

For the left branch, we process the feature $\mathbf{X}^{(L)} \in \mathbb{R}^{H \times W \times \frac{C}{2}}$ to capture local information via stacked convolutional layers. The equation is defined as follows:
\begin{equation}
	\mathbf{Y}_{m}^{(L)} = \text{Conv}(\text{ReLU}(\text{Conv}(\mathbf{X}^{(L)}))),
\end{equation}
\begin{equation}
    \mathbf{Y^{(L)}} = \text{PWConv}(\text{ReLU}(\mathbf{Y}_{m}^{(L)})).
\end{equation}

In the right branch, the input feature map $\mathbf{X}^{(R)} \in \mathbb{R}^{H \times W \times \frac{C}{2}}$ is first processed by Layer Normalization ($\text{LN}$) for normalization, and then expanded in the channel dimension via a linear transformation. Next, the feature map undergoes depthwise separable convolution for spatial feature extraction, followed by a nonlinear activation function. Subsequently, the feature map is scanned by SS2D in a direction-aware manner. Ultimately, these outputs are processed by a linear transformation and activation function to generate the final global feature representation. The equation is defined as follows:
\begin{equation}
	\mathbf{Y}_{0}^{(R)} = \text{LN}(\mathbf{X}^{(R)}), 
\end{equation}
\begin{equation}
	\mathbf{Y}_{1}^{(R)} = \text{SS2D}(\text{SiLU}(\text{DWConv}(\text{Linear}(\mathbf{Y}_{0}^{(R)}))))),
\end{equation}
\begin{equation}
	\mathbf{Y}_{2}^{(R)} = \text{LN}(\mathbf{Y}_{1}^{(R)}), 
\end{equation}
\begin{equation}
	\mathbf{Y}_{3}^{(R)} = \text{SiLU}(\text{Linear}(\mathbf{Y}_{0}^{(R)})),
\end{equation}
\begin{equation}
	\mathbf{Y}^{(R)} = \text{Linear}(\mathbf{Y_{2}^{(R)}} \odot \mathbf{Y}_{3}^{(R)}).
\end{equation}

Among them, for SS2D, the feature $\mathbf{S}$ is then fed into the SS2D,  which performs direction-aware selective scanning along four directions:  Vertical Forward (V-FWD), Horizontal Forward (H-FWD), Vertical Backward  (V-BWD), and Horizontal Backward (H-BWD). Each scanning  direction converts the 2D feature map into a 1D sequence by traversing  pixels in the respective order (e.g., top-to-bottom for V-FWD, left-to-right  for H-FWD). Each of these directions is processed separately as a feature  sequence $\mathbf{S}_{i}$ using the $\text{S6}$ block. Subsequently, the four directional outputs $\mathbf{S}'_{i}$ are merged via element-wise summation. The equation is defined as follows:
\begin{equation}
	\mathbf{S}'_{i} = \text{S6}(\text{expand}(\mathbf{S_i})), 
\end{equation}
\begin{equation}
	\mathbf{Y}_{\text{SS2D}} = \text{Merge}(\mathbf{S}'_{i}), \quad i \in \{1, 2, 3, 4\},
\end{equation}
where $\text{expand}(\mathbf{S_i})$ denotes unfolding the 2D feature map $\mathbf{S}_{i}$ into a 1D sequence along a specified direction.

Meanwhile, for the $\text{S6}$ block, the equation is defined as follows:
\begin{equation}
	\mathbf{\bar{A}} = e^{\Delta \mathbf{A}}, \quad \mathbf{\bar{B}} = \Delta \mathbf{B}, 
\end{equation}
\begin{equation}	
	\mathbf{h}_{t} = \mathbf{\bar{A}} \mathbf{h}_{t-1} + \mathbf{\bar{B}} \mathbf{x}_{t},
\end{equation}
\begin{equation}
	\mathbf{y}_{t} = \mathbf{C} \mathbf{h}_{t} + \mathbf{D} \mathbf{x}_{t}, 
\end{equation}
\begin{equation}
	\mathbf{y} = [\mathbf{y}_{1}, \mathbf{y}_{2}, \ldots, \mathbf{y}_{L}],
\end{equation}
where matrices $\mathbf{A} \in \mathbb{R}^{N \times N}$ (state dynamics), $\mathbf{B} \in \mathbb{R}^{N \times D}$ (input projection), $\mathbf{C} \in \mathbb{R}^{D \times N}$ (output projection), and $\mathbf{D} \in \mathbb{R}^{D}$ (residual connection) parameterize the respective operations, where $D$ denotes the channel dimension and $N$ denotes the state space dimension. $\mathbf{B}$ and $\mathbf{C}$ are 	input-dependent parameters computed via linear projections. Discretization via trainable $\Delta$ transforms $(\mathbf{A}, \mathbf{B})$ to $(\bar{\mathbf{A}}, \bar{\mathbf{B}})$. The model processes hidden states $\mathbf{h}_t \in \mathbb{R}^{N}$ to generate per-step outputs $\mathbf{y}_t \in \mathbb{R}^{D}$ and final output $\mathbf{y}$.

Furthermore, to enhance the importance of each channel in the feature representation, attention sharing is incorporated to emphasize critical information from each channel. Specifically, for each branch input $\mathbf{X}^{(i)} \in \mathbb{R}^{H \times W \times \frac{C}{2}}$ (where $i$ denotes the left or right branch), Global Average Pooling ($\text{G-AP}$) and Global
Max Pooling ($\text{G-MP}$) are applied separately to obtain the global features. The equation is defined as follows:
\begin{equation}
	\mathbf{G}_{avg} = \text{G-AP}(\mathbf{X}^{(i)}),
\end{equation}
\begin{equation}
	\mathbf{G}_{max} = \text{G-MP}(\mathbf{X}^{(i)}).
\end{equation}

Then, a shared convolutional network processes these two pooled results for attention weight computation. These processed features are combined through a Sigmoid activation function, yielding the final attention map $\mathbf{A}$. The equation is defined as follows:
\begin{equation}
	\mathbf{A}^{(i)}  = \sigma(\text{Conv}(\text{ReLU}(\text{Conv}(\mathbf{G}_{avg} + \mathbf{G}_{max})))).
\end{equation}

Finally, after enhancing the features of the left and right branches separately, they are merged and passed through a channel shuffle operation to improve inter-channel information interaction. The result is added element-wise to the original input features. Ultimately, it achieves effective fusion with the primary branch, thereby enabling joint modeling of global information and local details in the feature map. The equation is defined as follows:
\begin{equation}
	\mathbf{Y}_{c} = \text{Concat}(\mathbf{Y^{(L)}} \odot \mathbf{A}^{(L)}, \mathbf{Y^{(R)}} \odot \mathbf{A}^{(R)}), 
\end{equation}
\begin{equation}
	\mathbf{Y}_{\text{out}} =\text{Shuffle}(\mathbf{Y}_{c}) + \mathbf{X},
\end{equation}
where $\text{Shuffle}(\cdot)$ denotes the channel shuffle operation.
\begin{figure}[!t]
	\centering
	\includegraphics[width=0.5\textwidth]{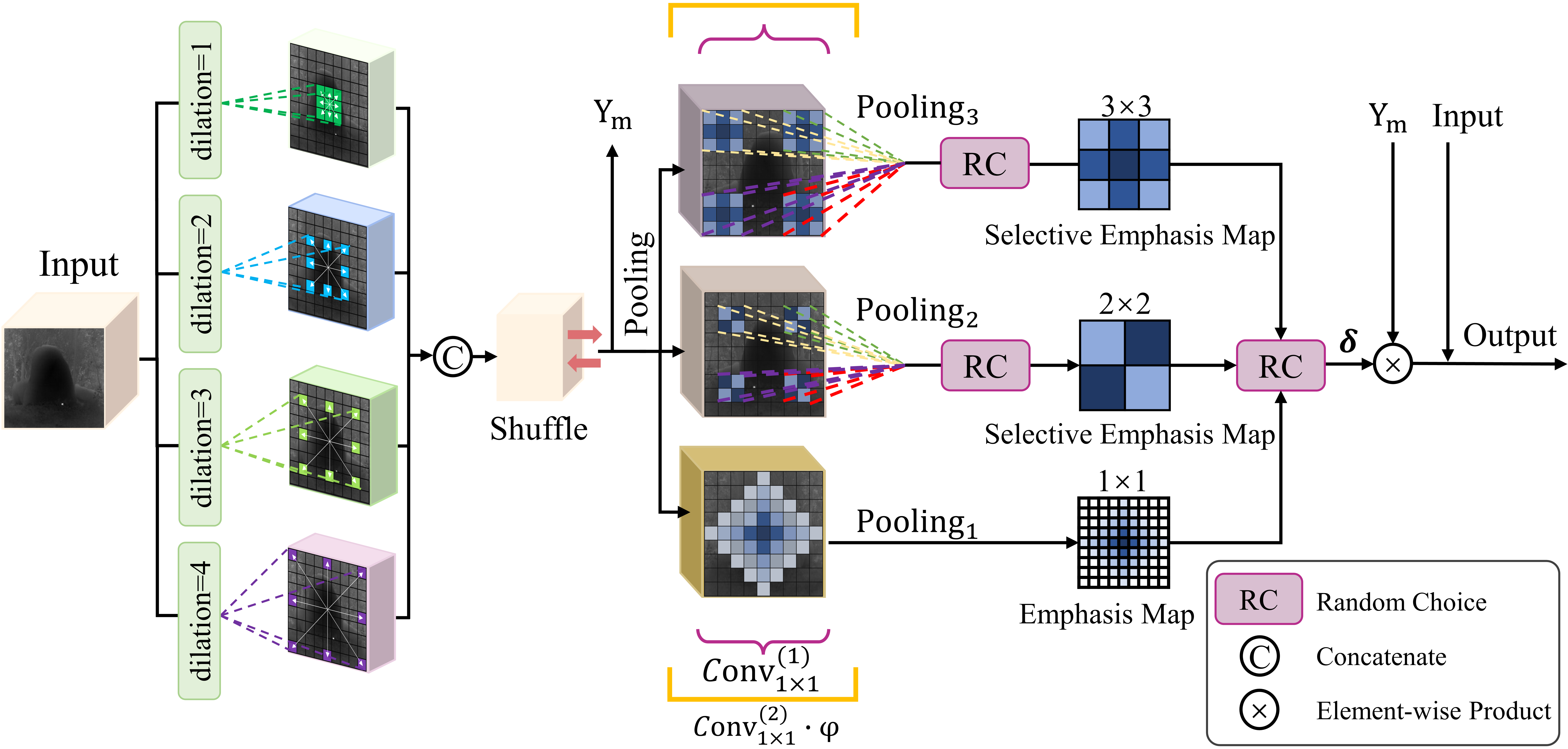}
	\caption{This figure illustrates the architecture of the LaSEA module. The module adopts cross-scale feature extraction and random pooling sampling strategies to provide more discriminative and robust target semantic feature representations, thereby enhancing the capability of guiding the decoder in fusion.}
	\label{Fig.3}
\end{figure}
\subsection{Latent-aware Semantic Extraction and Aggregation} 
Due to the inherent characteristics of infrared images, the target resolution becomes smaller in the deep features of the network, that is, the saliency of target semantics decreases in deeper layers. As a result, the network fails to provide the decoder with sufficient and accurate guidance for feature fusion. To address this challenge, we incorporate a LaSEA module (as shown in \text{Fig. \ref{Fig.3}}) into the bottleneck layer of the network. The LaSEA module employs a random pooling sampling strategy, randomly selecting from multiple pooling resolutions during training. Using a fixed pooling strategy causes the network to overfit to scale-specific feature responses, limiting feature diversity. In contrast, random selection across different pooling scales acts as a regularizer, enabling the network to learn robust discriminative features that remain stable across scales rather than overfitting to noise responses, thereby effectively suppressing background noise interference. Specifically, $\mathbf{X} \in \mathbb{R}^{H \times W \times C}$ is first passed through four spatial convolutions with different dilation rates to extract cross-scale features, which are then concatenated channel-wise to generate the integrated representation. The equation is defined as follows:
\begin{equation}
	\mathbf{Y}_{c} = \text{Concat}(\text{Conv}_{d_i}(\mathbf{X})), \quad i \in \{1,2,3,4\},
\end{equation}
where $d_i$ denotes the different dilation rates. 

To enhance the interaction between channels, the fused feature map first undergoes the channel shuffle operation. Then, during the training phase, we introduce a stochastic regularization mechanism: the system randomly selects one branch from the three candidate spatial pooling strategies $ \text{Pooling}_k(\cdot)$, corresponding to three pooling sizes: \( 3 \times 3 \), \( 2 \times 2 \), and \( 1 \times 1 \), and for pooling outputs larger than $1 \times 1$, randomly selects one spatial position from the pooled result. This randomization prevents overfitting to scale-specific feature responses, analogous to Dropout that randomly drops neurons during training. Gradients backpropagate only through the selected branch in each forward pass. The selected spatial feature is then used to generate the attention weights \( \boldsymbol{\delta} \) for subsequent feature enhancement. The equation is defined as follows:
\begin{equation}
	\mathbf{Y}_{m} = \text{Shuffle} \left( \mathbf{Y}_{c} \right),
\end{equation}
\begin{equation}
	\mathbf{v}_k = \text{Sample}_{\text{rand}}(\text{Pooling}_k(\mathbf{Y_{m}})), \quad k \in \{1, 2, 3\},
\end{equation}
\begin{equation}
	\boldsymbol{\delta} = \sigma(\text{Conv}_{1\times1}^{(2)}(\varphi(\text{Conv}_{1\times1}^{(1)}(\mathbf{v}_k)))),
\end{equation}
where $\text{Sample}_{\text{rand}}(\cdot)$ denotes random spatial 	sampling that selects one position from the pooled features during training, or global average pooling during inference; $\mathbf{v}_k$ is the resulting spatial feature vector; $\text{Conv}_{1\times1}^{(1)}$ and $\text{Conv}_{1\times1}^{(2)}$ denote two $1\times1$ convolutional operations with learnable parameters; $\varphi(\cdot)$ represents the ReLU activation function; $\sigma(\cdot)$ is the Sigmoid function; and $\boldsymbol{\delta}$ indicates the learned spatial attention weights.

Finally, the intermediate feature $\mathbf{Y}_{m}$ is multiplied element-wise by $\boldsymbol{\delta}$, and the result is added back to $\mathbf{X}$ pixel-wise to obtain the final output. The equation is defined as follows:
\begin{equation}
	\mathbf{Y}_{\text{out}} = \mathbf{X} + \left( \mathbf{Y}_{m} \odot \boldsymbol{\delta} \right).
\end{equation}

During the inference phase, a deterministic $1 \times 1$ pooling is employed to improve efficiency and stability. This training-inference discrepancy is similar to the Dropout principle: the stochastic training process encourages the network to learn robust features that generalize well across different pooling scales, while the deterministic inference maintains these advantages without introducing randomness.
\section{Experiments}
This section first describes the experimental setup, then presents quantitative and qualitative results. Subsequently, an ablation study is conducted to rigorously evaluate the effectiveness of the proposed network, and finally, visual analysis is performed using heatmaps.
\begin{table*}[!t]
	\centering
	\caption{Quantitative comparison with advanced methods on IRSTD-1K, NUAA-SIRST, and SIRST-Aug datasets using \textit{IoU}~(\%), \textit{Pd}~(\%), and \textit{Fa}~($\times 10^{-6}$). Bold blue indicates the best CNN method performance, bold red denotes the top hybrid method performance.}
	\label{tab:1}
	\renewcommand{\arraystretch}{1.2}
	\resizebox{\textwidth}{!}{%
		\begin{tabular}{l|c|c|ccc|ccc|ccc}
			\toprule
			\multirow{3}{*}{\centering\textbf{Methods}}  &\multirow{3}{*}{\centering\textbf{Type}} & \multirow{3}{*}{\centering\textbf{Publication}}
			& \multicolumn{3}{c|}{\textbf{IRSTD-1K}} 
			& \multicolumn{3}{c|}{\textbf{NUAA-SIRST}} 
			& \multicolumn{3}{c}{\textbf{SIRST-Aug}} \\
			\cmidrule(lr){4-6} \cmidrule(lr){7-9} \cmidrule(lr){10-12}
			& & & IoU$\uparrow$ & P$_d\uparrow$ & F$_a\downarrow$ 
			& IoU$\uparrow$ & P$_d\uparrow$ & F$_a\downarrow$
			& IoU$\uparrow$ & P$_d\uparrow$ & F$_a\downarrow$ \\
			\midrule
			\multicolumn{11}{@{}l}{\textit{Traditional Methods}} \\
			\midrule
			\rowcolor{purple!4!white}  
			Tophat \cite{tom1993morphology}        & Filtering & SPIE, 1993 & 10.06 & 75.11 & 1432.00 & 11.43 & 89.91 & 194.20 & 12.45 & 81.16 & 160.32 \\
			\rowcolor{purple!7!white}  
			MPCM \cite{wei2016multiscale}          & Local Contrast & PR, 2016 & 2.30  & 90.91 & 525.25 & 7.66 & 98.17  & 557.87  & 23.46 & 94.22 & 247.78  \\
			\rowcolor{purple!10!white}  
			IPI \cite{6595533}                     & Low Rank & TIP, 2013 & 27.92 & 81.37 & 16.18   & 18.06 & 90.83 & 35.33   & 19.05 & 75.79 & 101.30 \\
			\midrule
			\multicolumn{11}{@{}l}{\textit{Deep Learning Methods}} \\
			\midrule
			\rowcolor{gray!10}
			DNA-Net~\cite{9864119}                 & CNN & TIP, 2022 & 67.54 & 92.18 & 11.77   & 77.04 & 99.08 & 20.05   & 72.27 & \textbf{\textcolor{blue}{98.35}} & 51.85  \\
			UIU-Net~\cite{9989433}                 & CNN & TIP, 2022 & 65.59 & 88.44 & 14.73   & 76.83 & 98.17 & 11.71   & 70.64 & 96.56 & 63.47  \\
			\rowcolor{gray!10}
			RDIAN~\cite{10011452}                  & CNN & TGRS, 2023 & 63.40 & 92.86 & 11.54   & 74.08 & \textbf{\textcolor{blue}{100.00}} & 19.87 & 71.45 & 98.07 & 116.58 \\
			AGPC-Net~\cite{10024907}               & CNN & TAES, 2023 & 65.93 & 91.15 & \textbf{\textcolor{blue}{11.32}}   & \textbf{\textcolor{blue}{77.47}} & \textbf{\textcolor{blue}{100.00}} & \textbf{\textcolor{blue}{3.02}}  & 72.73 & 95.46 & 128.73 \\
			\rowcolor{gray!10}
			MSHNet~\cite{10658560}                 & CNN & CVPR, 2024 & 67.16 & \textbf{\textcolor{blue}{93.88}} & 15.03   & 73.65 & 99.08 & 19.16 & 72.64 & 96.56 & 91.55  \\
			HCF-Net~\cite{10687431}                & CNN & ICME, 2024 & 63.72 & 89.46 & 17.38   & 75.43 & 98.17 & 11.18 & 71.69 & 97.39 & \textbf{\textcolor{blue}{37.77}}  \\
			\rowcolor{gray!10}
			L2SKNet~\cite{10813615}               & CNN & TGRS, 2025 & 65.67 & 92.18 & 23.99   & 75.52 & 98.17 & 17.21 & 73.26 & 97.94 & 50.34  \\
			\rowcolor{gray!10}
			PConv~\cite{yang2025pinwheel}          & CNN & AAAI, 2025  & \textbf{\textcolor{blue}{67.58}} & \textbf{\textcolor{blue}{93.88}} & 12.22   & 75.98 & 98.17 & 4.26 & \textbf{\textcolor{blue}{74.04}} & 93.95 & 288.71  \\
			\midrule
			\rowcolor{orange!4!white} 
			ABC~\cite{10219645}                    & CNN-Transformer & ICME, 2023 & 65.13 & 88.43 & 14.96   & 77.59 & 98.17 & 6.74    & 72.86 & 98.76 & 56.64  \\
			\rowcolor{orange!7!white}  
			MTU-Net~\cite{10011449}                & CNN-Transformer & TGRS, 2023 & 65.44 & 87.75 & 16.40   & 77.46 & \textbf{\textcolor{red}{100.00}} & 5.68  & 71.59 & 96.42 & 37.57  \\
			\rowcolor{orange!10!white} 
			SCTransNet~\cite{10486932}             & CNN-Transformer & TGRS, 2024 & 66.07 & 92.52 & 15.79   & 76.18 & 99.08 & 26.97   & 73.79 & 98.21 & 46.67  \\
			\rowcolor{RoyalBlue!10}
			\textbf{DCCS-Det (Ours)}                  & CNN-Mamba & -- & \textbf{\textcolor{red}{69.64}} & \textbf{\textcolor{red}{95.58}} & \textbf{\textcolor{red}{10.48}} 
			& \textbf{\textcolor{red}{78.65}} & \textbf{\textcolor{red}{100.00}} & \textbf{\textcolor{red}{2.48}} 
			& \textbf{\textcolor{red}{75.57}} & \textbf{\textcolor{red}{98.90}} & \textbf{\textcolor{red}{33.46}} \\
			\bottomrule
		\end{tabular}%
	}
\end{table*}
\subsection{Experimental Settings}
\subsubsection{Datasets}
The proposed method is evaluated across three publicly released benchmark datasets: IRSTD-1K~\cite{zhang2022isnet}, NUAA-SIRST~\cite{dai2021asymmetric}, and SIRST-Aug~\cite{10024907}.

\textbf{IRSTD-1K} comprises 1,001 annotated images spanning six background scenarios—sea, river, field, mountain, city, and cloud. In accordance with the established protocol, 800 images contribute to training procedures and 201 images to validation procedures.

\textbf{NUAA-SIRST} comprises 427 annotated images across three background contexts—cloud, city, and sea. In accordance with the established protocol, 341 images contribute to training procedures and 86 images to validation procedures.

\textbf{SIRST-Aug} comprises 9,070 annotated images derived from real infrared imagery and augmented through operations such as rotation, cropping, and scaling, encompassing diverse scene types. In accordance with the established protocol, 8,525 images contribute to training procedures and 545 images to validation procedures.

\subsubsection{Implementation Details}

Experimental setup utilized an NVIDIA RTX 4090 GPU (24 GB memory). The PyTorch framework was employed for implementation. Data augmentation strategies included random flipping and rotation. Image dimensions were standardized to 256×256 pixels. Training configuration involved the AdaGrad optimizer over 400 epochs, using a 0.05 learning rate and batches of 4 samples. These settings remained consistent across all three datasets. 

\subsubsection{Evaluation Metrics}
We employ three commonly used metrics to quantify detection performance at different evaluation levels: the pixel-level metrics Intersection over Union ($IoU$) and False Alarm Rate ($F_a$), and the instance-level metric Probability of Detection ($P_d$). Specifically, $F_a$ and $P_d$ measure the detector's tendency toward false alarms and its recall capability, respectively, while $IoU$ serves as a representative pixel-level metric that evaluates the degree of overlap between the predicted and ground-truth regions, defined as the ratio of their intersection area to their union area. The equation is defined as follows:
\begin{equation}
	IoU = \frac{A_i}{A_u} = 
	\frac{\sum_{i=1}^{N} TP[i]}
	{\sum_{i=1}^{N} \big(T[i] + P[i] - TP[i]\big)},
\end{equation}
\begin{equation}
	P_d = \frac{N_{pred}}{N_{all}},
\end{equation}
\begin{equation}
	F_a = \frac{P_{false}}{P_{all}},
\end{equation}
where $A_i$ and $A_u$ denote the intersection and union areas between the predicted and ground-truth regions. $N$ is the total number of pixels in the image. $TP[\cdot]$ denotes the number of true positive pixels correctly predicted as targets, while $T[\cdot]$ and $P[\cdot]$ represent the ground-truth and predicted target masks, respectively. $N_{pred}$ is the number of correctly detected target instances, and $N_{all}$ is the total number of targets. Consistent with~\cite{9864119}, a target instance is considered correctly detected for $P_d$ when the centroid distance between the ground-truth and the nearest predicted instance is within 3 pixels. $P_{false}$ refers to the number of background pixels incorrectly predicted as targets, and $P_{all}$ represents the total number of pixels in the image. Similarly, all $F_a$ values in this paper are scaled by $10^{6}$ to represent false alarms per megapixel.

Additionally, model performance is assessed through Receiver Operating Characteristic (ROC) curves analysis, which illustrates the relationship between pixel-level True Positive Rate ($TPR$) and False Positive Rate ($FPR$) across different threshold values. The equation is defined as follows:
\begin{equation}
	TPR = \frac{\sum_{i=1}^{N} TP[i]}{\sum_{i=1}^{N} \big(TP[i] + FN[i]\big)},
\end{equation}
\begin{equation}
	FPR = \frac{\sum_{i=1}^{N} FP[i]}{\sum_{i=1}^{N} \big(TN[i] + FP[i]\big)},
\end{equation}
where $TN[\cdot]$, $FP[\cdot]$, and $FN[\cdot]$ denote the number of true negative pixels, false positive pixels, and false negative pixels.

\begin{table*}[!t]
	\centering
	\caption{Efficiency comparisons on the IRSTD-1K benchmark across Params (M), FLOPs (G), Time (ms), FPS, and \textit{IoU}~(\%). Bold blue indicates the best CNN method performance, bold red denotes the top hybrid method performance.}
	\label{tab:2}
	\renewcommand{\arraystretch}{1.2}
	\setlength{\tabcolsep}{10pt} 
	\footnotesize 
	\resizebox{0.95\textwidth}{!}{
		\begin{tabular}{l|c|c|cccc|c}
			\toprule
			\multirow{3}{*}{\centering\textbf{Methods}} & \multirow{3}{*}{\centering\textbf{Type}} & \multirow{3}{*}{\centering\textbf{Publication}}
			& \multicolumn{4}{c|}{\textbf{Efficiency}} 
			& \multirow{3}{*}{\centering\textbf{IoU$\uparrow$}} \\
			\cmidrule(lr){4-7}
			& & & \textbf{Params (M)} & \textbf{FLOPs (G)} & \textbf{Time (ms)} & \textbf{FPS (f/s)} & \\
			\midrule
			\rowcolor{gray!10}
			DNA-Net~\cite{9864119}      & CNN         & TIP,2022    & 4.70  & 14.26 & 44.07 & 22.69  & 67.54 \\
			UIU-Net~\cite{9989433}      & CNN         & TIP,2022    & 50.54 & 54.43 & 41.95 & 23.84  & 65.59 \\
			\rowcolor{gray!10}
			RDIAN~\cite{10011452}        & CNN         & TGRS,2023   & \textbf{\textcolor{blue}{0.22}}  & \textbf{\textcolor{blue}{3.74}}  & 14.94 & 66.93  & 63.40 \\
			AGPC-Net~\cite{10024907}    & CNN         & TAES,2023   & 12.36 & 86.36 & 67.28 & 14.86  & 65.93 \\
			\rowcolor{gray!10}
			MSHNet~\cite{10658560}      & CNN         & CVPR,2024   & 4.07  & 6.11  & 17.09 & 58.52  & 67.16 \\
			HCF-Net~\cite{10687431}        & CNN         & ICME,2024   & 14.40 & 23.28 & 45.49 & 21.98  & 63.72 \\
			\rowcolor{gray!10}
			L2SKNet~\cite{10813615}     & CNN         & TGRS,2025   & 0.90  & 6.91  & \textbf{\textcolor{blue}{12.85}} & \textbf{\textcolor{blue}{77.82}}  & 65.67 \\
			PConv~\cite{yang2025pinwheel}        & CNN         & AAAI,2025   & 4.04  & 6.05  & 16.73 & 59.77  & \textbf{\textcolor{blue}{67.58}} \\
			\midrule
			\rowcolor{orange!4!white}
			ABC~\cite{10219645}              & CNN-Transformer & ICME,2023   & 73.51 & 83.13 & 41.63 & 24.02  & 65.13 \\
			\rowcolor{orange!7!white}
			MTU-Net~\cite{10011449}       & CNN-Transformer & TGRS,2023   & 28.61 & 23.52 & 36.54 & 27.37  & 65.44 \\
			\rowcolor{orange!10!white}
			SCTransNet~\cite{10486932}   & CNN-Transformer & TGRS,2024   & 11.19 & 10.12 & 47.67 & 20.98  & 66.07 \\
			\rowcolor{RoyalBlue!10}
			\textbf{DCCS-Det (Ours)} & CNN-Mamba & -- & \textbf{\textcolor{red}{7.09}} & \textbf{\textcolor{red}{7.97}} & \textbf{\textcolor{red}{21.09}} & \textbf{\textcolor{red}{47.42}} & \textbf{\textcolor{red}{69.64}} \\
			\bottomrule
		\end{tabular}
	}
\end{table*}
\begin{figure*}[]
	\centering
	\includegraphics[width=\textwidth]{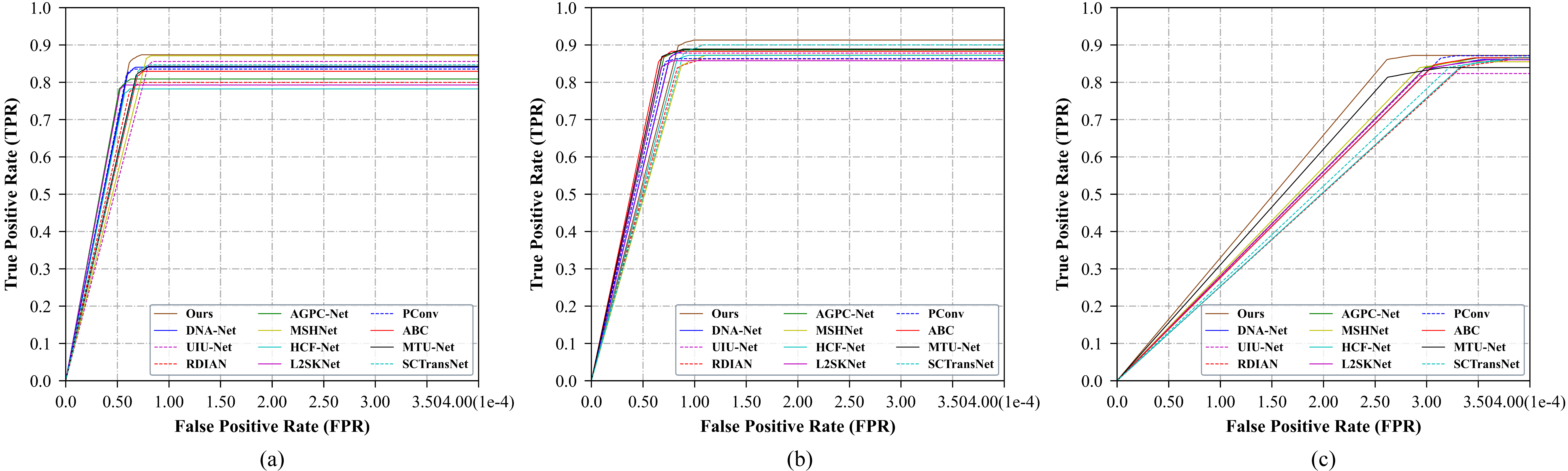}
	\caption{Comparison of ROC curves across various approaches on (a) IRSTD-1K, (b) NUAA-SIRST, and (c) SIRST-Aug datasets. The horizontal axis represents FPR while the vertical axis represents TPR. Curves approaching the top-left corner indicate superior detection performance.}
	\label{Fig.11}
\end{figure*}
\subsubsection{Baseline Methods}
The proposed method is benchmarked against three traditional approaches and eleven recent deep learning approaches. The conventional methods include Tophat~\cite{tom1993morphology}, MPCM~\cite{wei2016multiscale}, and IPI~\cite{6595533}. The deep learning counterparts consist of nine CNN-based models, including DNA-Net~\cite{9864119}, UIU-Net~\cite{9989433}, RDIAN~\cite{10011452}, AGPC-Net~\cite{10024907}, MSHNet~\cite{10658560}, HCF-Net~\cite{10687431}, L2SKNet~\cite{10813615} and PConv~\cite{yang2025pinwheel}, as well as three Transformer-based hybrid architectures: ABC~\cite{10219645}, MTU-Net~\cite{10011449}, and SCTransNet~\cite{10486932}. To ensure fair comparison, DCCS-Det and all comparative deep learning models were retrained at unified $256 \times 256$ resolution with the same data augmentation, while each model preserves its original network architecture and training settings from the corresponding publication.

\begin{figure*}[!t]
	\centering
	\includegraphics[width=\textwidth]{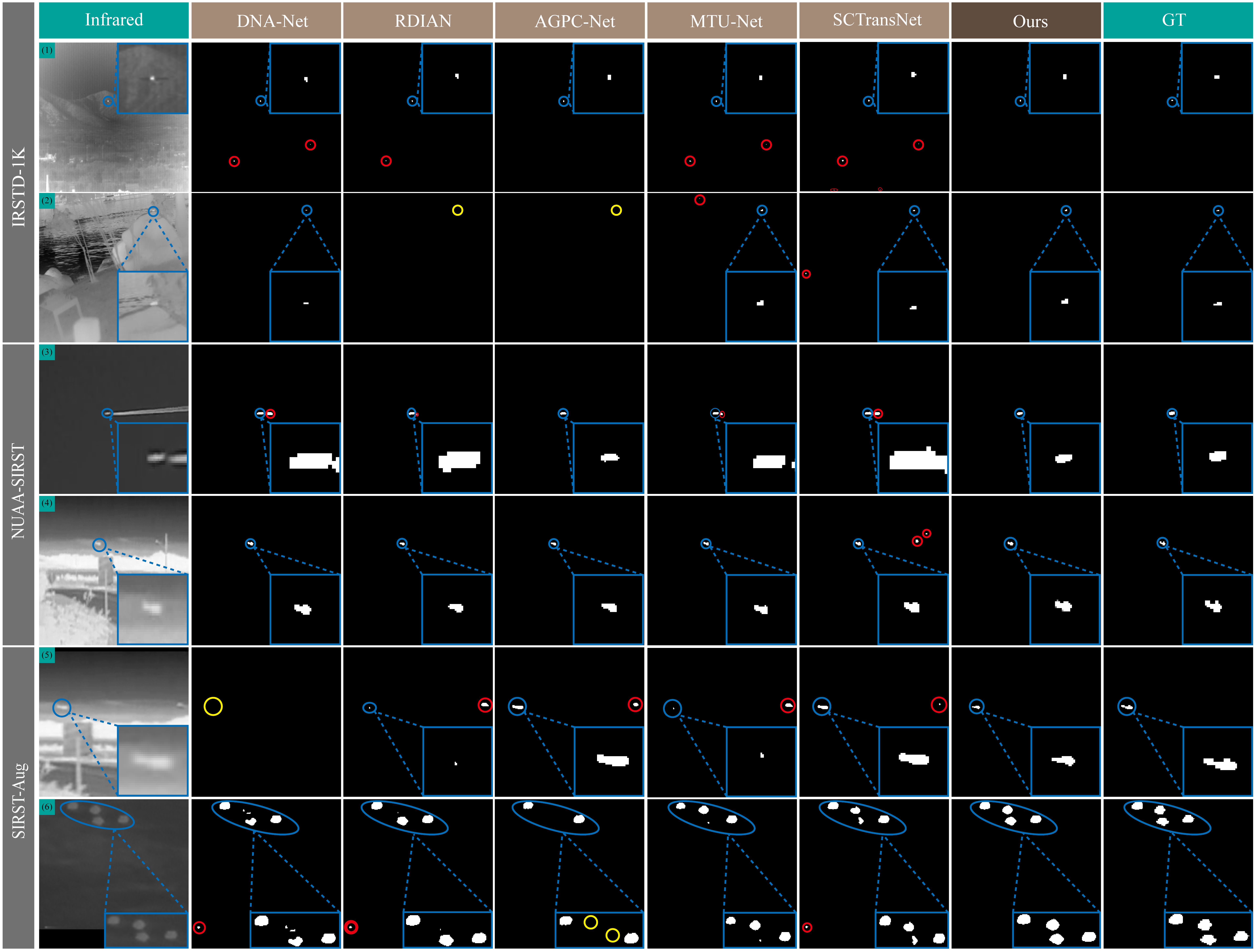}
	\caption{Visual results of different IRSTD methods on the IRSTD-1K, NUAA-SIRST, and SIRST-Aug datasets. Blue, yellow, and red circles respectively denote correctly detected targets, missed targets, and false alarms.}
	\label{Fig.4}
\end{figure*}

\subsection{Quantitative Results}
To validate the effectiveness and superiority of the proposed DCCS-Det method in IRSTD tasks, we conduct comprehensive performance comparisons with various existing methods. Table~\ref{tab:1} presents quantitative comparisons of several representative target detection methods on three widely used IRSTD benchmark datasets (IRSTD-1K, NUAA-SIRST, and SIRST-Aug). The results indicate that traditional methods significantly lag behind deep learning models in terms of performance. Among deep learning-based methods, pure CNN architectures (such as RDIAN, AGPC-Net, etc.) demonstrate excellent performance on certain datasets, with AGPC-Net achieving a Pd value of 100.00 on NUAA-SIRST, but its Fa still reaches 3.02, while on SIRST-Aug its Fa is as high as 128.73. CNN-Transformer hybrid architecture methods, by combining the advantages of both architectures, show relatively limited overall performance. For instance, ABC achieves IoU values of 65.13, 77.59, and 72.86 across the three datasets, MTU-Net achieves 65.44, 77.46, and 71.59, and SCTransNet achieves 66.07, 76.18, and 73.79. In contrast, our proposed DCCS-Det adopts a CNN-Mamba architecture and achieves competitive performance across three key metrics, achieving IoU values of 69.64, 78.65, and 75.57, while maintaining low Fa values of 10.48, 2.48, and 33.46, demonstrating the significant advantages of the CNN-Mamba architecture in IRSTD tasks.

To demonstrate the computational efficiency and practical applicability of the proposed DCCS-Det method, Table~\ref{tab:2} presents a comparative analysis between the proposed DCCS-Det method and eleven representative deep learning-based target detection methods on the IRSTD-1K dataset. Among three different architecture types, pure CNN architecture methods exhibit significant performance variations, with lightweight networks such such as RDIAN with 0.22M parameters and 66.93 FPS demonstrating high computational efficiency with reasonable IoU values of 63.40,  while complex architectures such as UIU-Net with 50.54M parameters and 23.84 FPS show large parameter count and limited accuracy. CNN-Transformer hybrid architecture methods combine the advantages of both architectures. However, they have a relatively large number of parameters and limited effectiveness, as ABC has the highest parameter count of 73.51M but achieves only 65.13 IoU. The CNN-Mamba architecture DCCS-Det method achieves optimal balance, obtaining the highest IoU value of 69.64 with moderate parameter count of 7.09M and FLOPs of 7.97G, while maintaining high inference speed of 47.42 FPS, demonstrating the significant advantages of the CNN-Mamba architecture in balancing detection accuracy and computational efficiency.

To further evaluate the detection capabilities and confirm the superiority of DCCS-Det across different scenarios, \text{Fig. \ref{Fig.11}}  shows the ROC curves of various methods on three public IRSTD datasets. The results demonstrate that our method achieves superior detection accuracy across all datasets, especially maintaining a high TPR under extremely low FPR conditions. The method significantly outperforms advanced methods such as AGPC-Net, and UIU-Net, exhibiting excellent cross-dataset generalization capability.

To rigorously evaluate detection performance under extreme conditions, we conduct targeted experiments on extremely low signal-to-noise ratio (SNR) image subsets. Following standard protocols in the literature~\cite{10924406}, we construct three nested subsets from the IRSTD-1K test set based on SNR thresholds, where lower SNR corresponds to stronger background interference and weaker target signals. As shown in Table~\ref{tab:low_snr}, DCCS-Det demonstrates superior performance across all SNR ranges. Compared to other methods, DCCS-Det exhibits stronger robustness under extreme noise conditions, validating the superiority of our design.

\begin{table}[htbp]
	\renewcommand{\arraystretch}{1.3}
	\setlength{\tabcolsep}{6pt} 
	\centering
	\caption{Robustness evaluation on extremely low SNR subsets of IRSTD-1K (\textit{IoU}~\%).}
	\label{tab:low_snr}
	\resizebox{0.48\textwidth}{!}{%
		\begin{tabular}{l|c|c|c|c}
			\toprule
			\textbf{Method} &\textbf{Publication}& \textbf{SNR$<$3} & \textbf{SNR$<$4} & \textbf{SNR$<$5} \\
			\midrule
			DNA-Net~\cite{9864119}           & TIP,2022 & 52.83 & 59.06 & 61.58 \\
			\rowcolor{gray!5}
			UIU-Net~\cite{9989433}           & TIP,2022 & 51.32 & 57.23 & 59.00 \\
			AGPC-Net~\cite{10024907}         & TAES,2023 & 51.48 & 57.96 & 59.47 \\
			\rowcolor{gray!5}
			MSHNet~\cite{10658560}           & TGRS,2024 & 50.00 & 56.03 & 59.43 \\
			HCF-Net~\cite{10687431}          & ICME,2024 & 52.76 & 58.25 & 61.11 \\
			\rowcolor{gray!5}
			SCTransNet~\cite{10486932}       & TGRS,2024 & 50.08 & 56.38 & 59.38\\
			PConv~\cite{yang2025pinwheel}    & AAAI,2025 & 51.75 & 59.37 & 61.65 \\
			\midrule
			\rowcolor{RoyalBlue!10}
			\textbf{DCCS-Det (Ours)} & -- & \textbf{\textcolor{red}{53.36}} & \textbf{\textcolor{red}{60.49}} & \textbf{\textcolor{red}{62.95}} \\
			\bottomrule
		\end{tabular}%
	}
\end{table}
\begin{figure*}[!t]
	\centering
	\includegraphics[width=\textwidth]{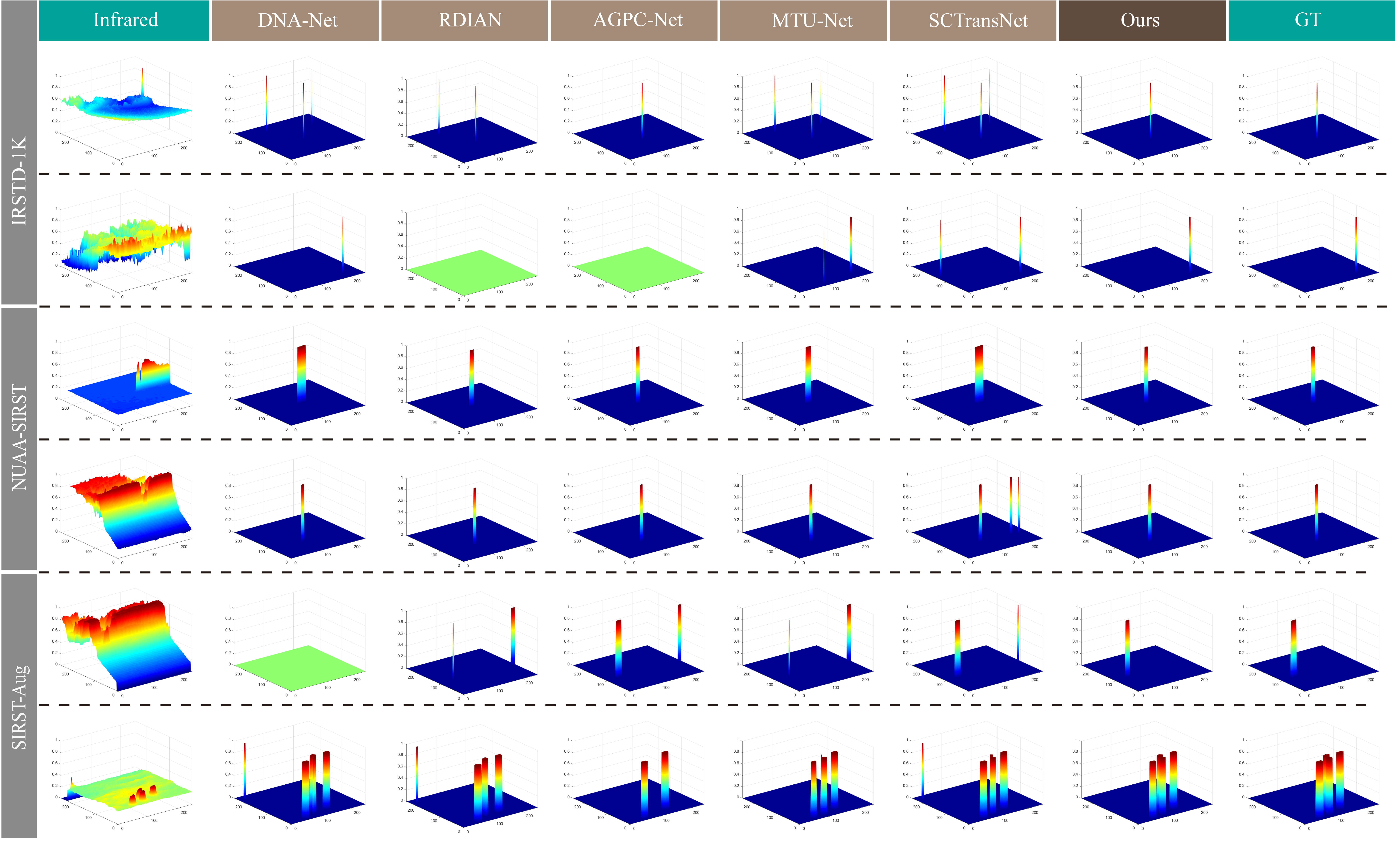}
	\caption{The 3D visualization results of different IRSTD methods on the IRSTD-1K, NUAA-SIRST, and SIRST-Aug datasets.}
	\label{Fig.12}
\end{figure*}

\subsection{Visual Comparisons}
The qualitative visual comparisons of our proposed DCCS-Det against five mainstream methods across three datasets are shown in \text{Figs.~\ref{Fig.4}, \ref{Fig.12}}. CNN-based methods, such as DNA-Net, RDIAN and AGPCNet, often exhibit false alarm and missed detections, especially in complex backgrounds or scenes with weak targets. For instance, in the first row, DNA-Net, and RDIAN show multiple false alarms; in the fifth row, RDIAN presents both false alarms, while DNA-Net misses targets. Transformer-based hybrid methods, such as MTU-Net and SCTransNet, though strong in global perception, still misidentify background areas as targets in scenes with significant interference or multiple targets. In contrast, DCCS-Det demonstrates superior detection performance across all datasets and scenarios. It not only accurately locates faint targets but also preserves their complete contours. For example, in the sixth row of the SIRST-Aug dataset, most methods fail to detect all targets or produce incorrect segmentations, while DCCS-Det successfully detects all targets with well-defined boundaries. This advantage is due to DCCS-Det's ability to jointly model global and local features of both targets and backgrounds with linear complexity, while enhancing deeper layers representations and effectively guiding decoder fusion, ultimately outperforming existing methods.

\begin{table}[h]
	\renewcommand{\arraystretch}{1.2}
	\centering
	\caption{Ablation study on the IRSTD-1K dataset evaluated by \textit{IoU}~(\%), \textit{Pd}~(\%), and \textit{Fa}~($\times 10^{-6}$).}
	\label{tab:3}
	\resizebox{0.48\textwidth}{!}{%
		\begin{tabular}{c|c|c|c|c|c|c}
			\toprule
			\textbf{No.} & \textbf{Baseline} & \textbf{DSE} & \textbf{LaSEA} & \textbf{IoU$\uparrow$} & \textbf{P$_d\uparrow$} & \textbf{F$_a\downarrow$} \\
			\midrule
			1 & \checkmark & $\times$    & $\times$    & 67.16 & 93.88 & 15.03 \\
			\rowcolor{gray!10}
			2 & \checkmark & \checkmark  & $\times$    & 68.05 & 95.24 & 13.21 \\
			3 & \checkmark & $\times$    & \checkmark  & 68.72 & 94.22 &  \textbf{\textcolor{red}{8.50}} \\
			\rowcolor{RoyalBlue!10}
			4 & \checkmark & \checkmark  & \checkmark  & \textbf{\textcolor{red}{69.64}} & \textbf{\textcolor{red}{95.58}} & 10.48 \\
			\bottomrule
		\end{tabular}%
	}
\end{table}

\begin{table}[h]
	\renewcommand{\arraystretch}{1.2}
	\footnotesize 
	\centering
	\caption{Ablation study on different connection fusion strategies 	between DSE Block and Res-Block in terms of \textit{IoU}~(\%), \textit{Pd}~(\%), and \textit{Fa}~($\times 10^{-6}$).}
	\label{tab:4}
	\resizebox{0.48\textwidth}{!}{%
		\begin{tabular}{lc|c|c|c}
			\toprule
			\textbf{Strategy} && \textbf{IoU$\uparrow$} & \textbf{P$_d\uparrow$} & \textbf{F$_a\downarrow$} \\
			\midrule
			Matrix Product Fusion && 67.11 & 94.22 & 13.89 \\
			\rowcolor{gray!10}
			Add && 66.48 & 94.56 & 15.33 \\
			Element-wise Mul && 66.41 & 93.54 & 16.32 \\
			\rowcolor{gray!10}
			EFC~\cite{10671587} && 66.37 & 94.90 & 20.88 \\
			LCA~\cite{yan2025hvi} && 66.61 & 90.82 & 21.92 \\
			\rowcolor{gray!10}
			Concat + SSA~\cite{li2025mair} && 66.82 & 93.54 & 14.98 \\
			Concat + CBAM~\cite{woo2018cbam} && 66.51 & 92.86 & 16.78 \\
			\rowcolor{gray!10}
			Concat + EMAttention~\cite{10096516} && 67.21 & 93.20 & 13.21 \\
			Concat + TripletAttention~\cite{9423300} && 67.68 & 93.54 & 14.68 \\
			\rowcolor{gray!10}
			Add + SSA~\cite{li2025mair} && 66.31 & 94.22 & 15.87 \\
			Add + CBAM~\cite{woo2018cbam} && 67.14  & 91.50 & 13.21 \\
			\rowcolor{gray!10}
			Add + EMAttention~\cite{10096516} && 66.51 & 91.16 & 15.11 \\
			Add + TripletAttention~\cite{9423300} && 67.30 & 93.20 & 13.51 \\
			\rowcolor{RoyalBlue!10}
			\textbf{Concat (Ours)} && \textbf{\textcolor{red}{68.05}} & \textbf{\textcolor{red}{95.24}} & \textbf{\textcolor{red}{13.21}} \\
			\bottomrule
		\end{tabular}%
	}
\end{table}
\subsection{Ablation Experiments}
\subsubsection{Effectiveness of Network Components}
To verify the effectiveness of the proposed DSE block and LaSEA module, a systematic evaluation was conducted on the IRST-D1K dataset. Four different model configurations were tested: without any modules (Baseline), with only DSE, with only LaSEA, and with both modules integrated. The results show that the Baseline model performed the worst. After integrating the DSE block, performance improved, indicating that the DSE block supplements and enhances global context information and local detail modeling, making the semantic response difference between targets and interference more distinct. When only the LaSEA module was added, performance further improved, especially in reducing false alarms. This suggests that the LaSEA module enhances the representation of target regions by extracting cross-scale feature extraction and random pooling sampling strategies, effectively guiding decoder fusion. Finally, when the DSE block and LaSEA modules were integrated, our proposed model demonstrated superior performance metrics. The detailed quantitative evaluations are shown in Table~\ref{tab:3}. Overall, the ablation study clearly validates the contributions of the DSE block and LaSEA module in improving detection accuracy, reducing false alarms, and enhancing the robustness of DCCS-Det. It outperforms all variants, highlighting the effectiveness and superiority of each network component.
\begin{figure*}[!t]
	\centering
	\includegraphics[width=\textwidth]{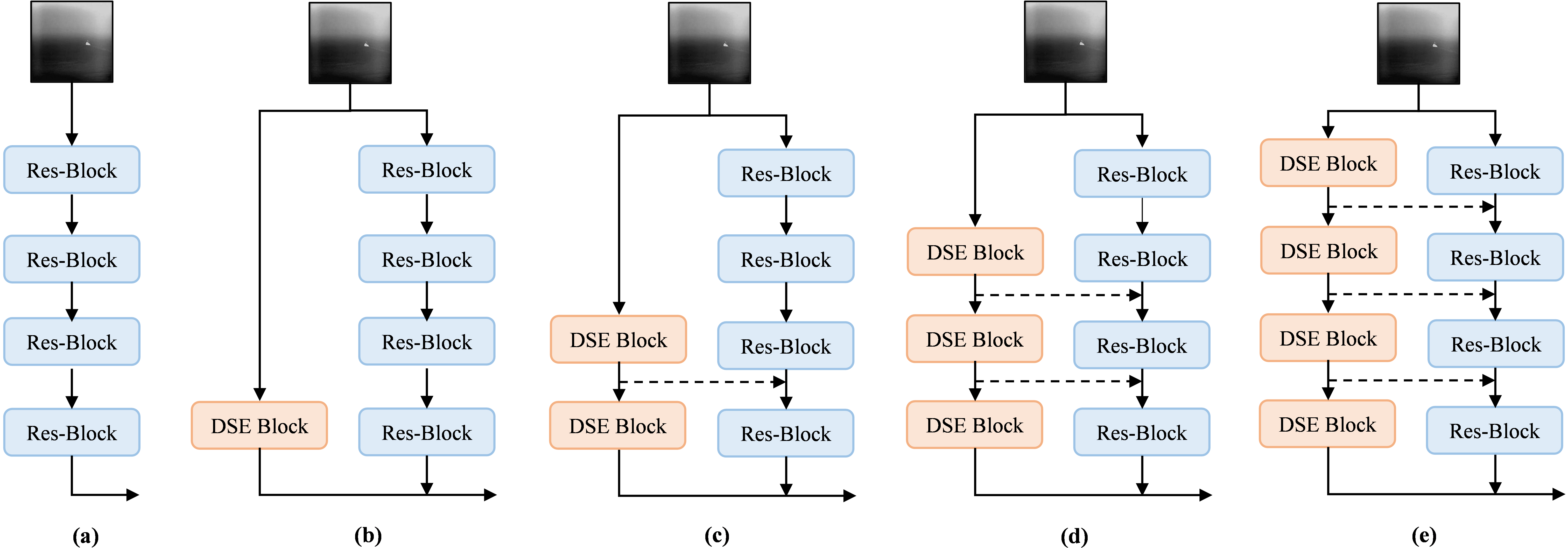}
	\caption{Structural illustration of four fusion strategies between DSE Block and Res-Block at different network depths: (a) Baseline, (b) Last layer only, (c) 2nd layer only, (d) 3 layers, and (e) All layers. The corresponding performance is reported in Table~\ref{tab:5}.}
	\label{Fig.7}
\end{figure*}
\subsubsection{Effect of Connection Methods on Model Performance}
To evaluate the impact of different connection fusion strategies between the DSE block and the primary Res-Block on model performance, we designed a series of ablation experiments, as shown in Table~\ref{tab:4}. The experiments covered several basic fusion methods, such as Add, Element-wise Mul, and Matrix Product Fusion, as well as several attention-based fusion strategies proposed in recent years, including SSA~\cite{li2025mair}, CBAM~\cite{woo2018cbam}, EMAttention~\cite{10096516} and TripletAttention~\cite{9423300}. We systematically compared the effectiveness of the above fusion methods based on the task of IRSTD. The experimental results show that traditional addition (Add) and element-wise multiplication (Element-wise Mul) methods achieved IoU values of 66.48 and 66.41. Although these methods are relatively simple to implement, they sacrifice the discriminability of features during the fusion process, resulting in a significant drop in model performance. Matrix Product Fusion retains structural information to some extent, but its overall performance is still inferior to feature concatenation (Concat). Among attention-based fusion methods, approaches such as Concat or Add combined with SSA, CBAM, EMAttention, and TripletAttention achieve improvements in certain evaluation metrics. However, their overall performance remains slightly lower than that of the pure Concat fusion strategy. This indicates that although the introduction of additional attention modules can enhance feature modeling capabilities, it may also increase network complexity and introduce noisy features, thereby negatively affecting the final performance. After comprehensive consideration, we ultimately adopted the Concat fusion strategy. This strategy demonstrates the best stability and effectiveness in our task, maximally preserving the completeness of dual-branch feature representations while effectively avoiding excessive compression and redundant interference.

\subsubsection{Impact of Fusion Hierarchies on Model Performance}
\begin{table}[!t]
	\renewcommand{\arraystretch}{1.2}
	\centering
	\setlength{\tabcolsep}{5.5pt} 
	\caption{Ablation study on fusion depth between DSE Block and Res-Block with Params (M), FLOPs (G), \textit{IoU}~(\%), \textit{Pd}~(\%), and \textit{Fa}~($\times 10^{-6}$).}
	\label{tab:5}
	\resizebox{0.48\textwidth}{!}{%
		\begin{tabular}{c|c|c|c|c|c}
			\toprule
			\textbf{Strategy} & \textbf{Params (M)} & \textbf{FLOPs (G)} & \textbf{IoU$\uparrow$} & \textbf{P$_d\uparrow$} & \textbf{F$_a\downarrow$} \\
			\midrule
			(a)                    & \textbf{\textcolor{red}{4.0655}} & \textbf{\textcolor{red}{6.1118}} & 67.16 & 93.88 & 15.03 \\
			\rowcolor{gray!10}
			(b)     & 4.4500 & 7.1960 & 66.28 & 94.90 & 15.49 \\
			(c)      & 4.4582 & 7.2295 & 66.54 & 94.56 & 17.31 \\
			\rowcolor{gray!10}
			(d)            & 4.4603 & 7.2631 & 67.33 & 93.20 & 13.66 \\
			\rowcolor{RoyalBlue!10}
			\textbf{(e)} & 4.4609 & 7.2967 & \textbf{\textcolor{red}{68.05}} & \textbf{\textcolor{red}{95.24}} & \textbf{\textcolor{red}{13.21}} \\
			\bottomrule
		\end{tabular}
	}
\end{table}

To evaluate the impact of different fusion strategies between the DSE Block and the Res-Block on model performance, four fusion schemes were designed, as shown in \text{Fig.~\ref{Fig.7}}, specifically: (a) Baseline, (b) one layer fusion, (c) two layers fusion, (d) three layers fusion, and (e) all layers fusion. Table~\ref{tab:5} presents the quantitative results. Among these four fusion strategies, all layers fusion (e) achieved the best performance, indicating that the all layers fusion strategy can better model the target context and local details. In contrast, reducing the number of fusion layers (as in strategies b, c, and d) leads to varying degrees of accuracy degradation and an increase in false alarm rates, thereby verifying the effectiveness and necessity of the all layers fusion strategy.

\subsubsection{Ablation study on different depth settings of the DSE Block}

\begin{table}[!t]
	\renewcommand{\arraystretch}{1.2}
	\centering
	\setlength{\tabcolsep}{5.5pt} 
	\caption{Ablation study on different depth settings of the DSE Block, quantified by Params (M), FLOPs (G), \textit{IoU}~(\%), \textit{Pd}~(\%), and \textit{Fa}~($\times 10^{-6}$).}
	\label{tab:6}
	\resizebox{0.48\textwidth}{!}{%
		\begin{tabular}{c|c|c|c|c|c }
			\toprule
			\textbf{Depths} & \textbf{Params (M)} & \textbf{FLOPs (G)} & \textbf{IoU$\uparrow$} & \textbf{P$_d\uparrow$} & \textbf{F$_a\downarrow$} \\
			\midrule
			{[1, 1, 1, 1]} & 4.31 & \textbf{\textcolor{red}{6.83}}  & 67.03 & \textbf{\textcolor{red}{95.24}} & 16.78 \\
			\rowcolor{RoyalBlue!8}
			\textbf{[2, 2, 2, 2]} & 4.46 & 7.30  & \textbf{\textcolor{red}{68.05}} & \textbf{\textcolor{red}{95.24}} & \textbf{\textcolor{red}{13.21}} \\
			{[0, 2, 4, 2]} & 4.52 & 7.26  & 67.64 & 94.90 & 17.76 \\
			\rowcolor{gray!10}
			{[2, 4, 2, 0]} & \textbf{\textcolor{red}{4.25}} & 7.32 & 67.94 & 91.84 & 13.66 \\
			{[4, 2, 2, 2]} & 4.46 & 7.56  & 67.44 & 94.22 & 23.76 \\
			\rowcolor{gray!10}
			{[2, 2, 2, 4]} & 4.69 & 7.51 & 67.61 & 94.56 & 13.59 \\
			\bottomrule
		\end{tabular}
	}
\end{table}

To evaluate the impact of the depths parameter setting of SS2D in the DSE block on model performance, multiple ablation experiments were conducted. Table~\ref{tab:6} presents these findings. The results indicate that when the depth of each stage is uniformly set to [2, 2, 2, 2], it can effectively enhance the collaborative modeling capability between the DSE Block and the Res-Block, improve the discrimination between targets and interference, and thereby enable the model to achieve the best performance across various metrics, indicating that a moderate depth configuration is reasonable and effective. In contrast, other asymmetric or single-stage depth configurations, such as [0, 2, 4, 2] and also [2, 4, 2, 0], achieve comparable results in some metrics, but overall performance slightly degrades. Therefore, a balanced and moderate depth setting not only enhances the DSE Block's effectiveness, but also forms a more symmetric and cooperative feature fusion modeling structure with the primary branch, while achieving a good balance among accuracy, parameter count, and computational complexity.
\subsubsection{Ablation Study on Attention Placement in the DSE Block}

\begin{figure}[!t]
	\centering
	\includegraphics[width=0.5\textwidth]{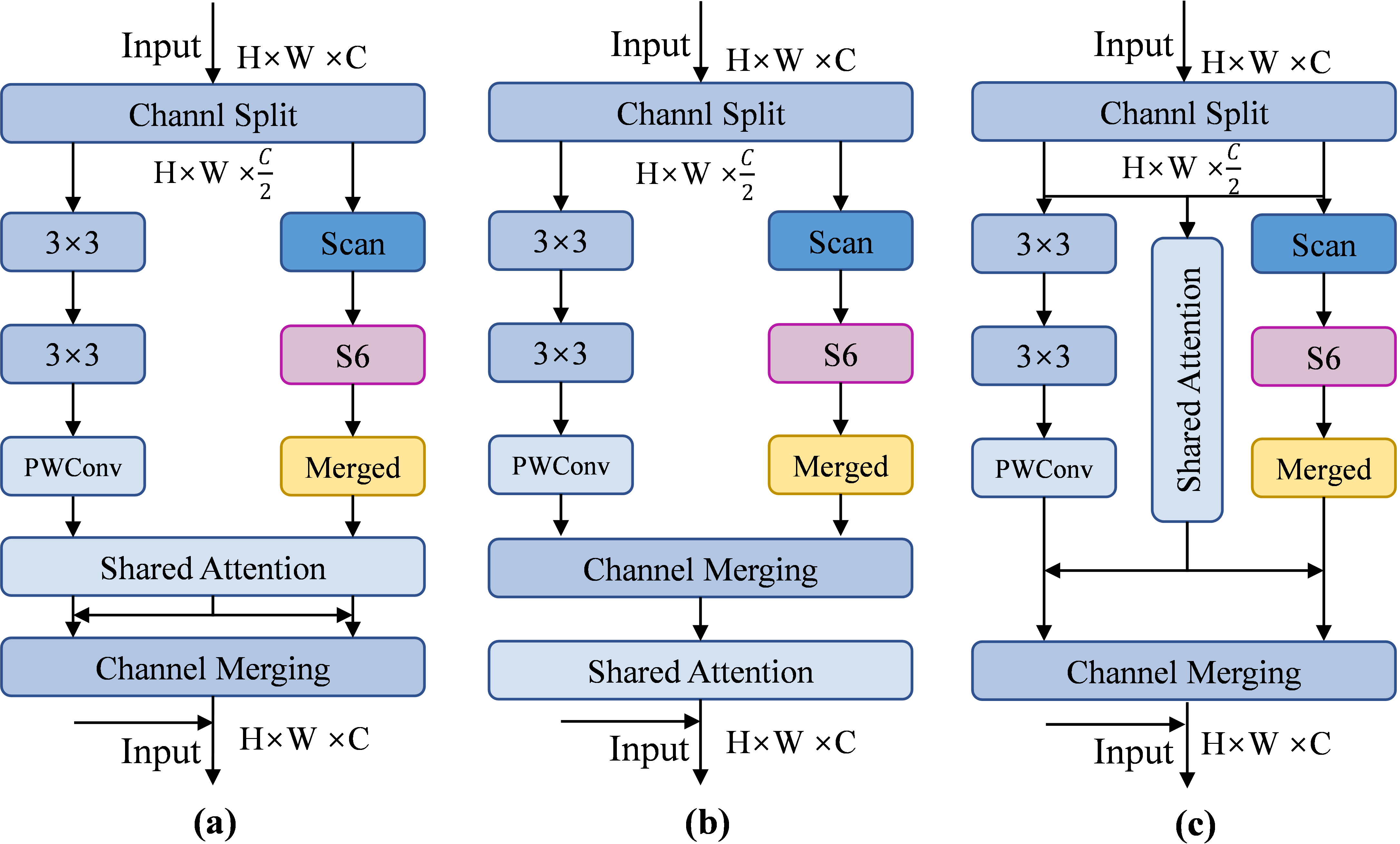}
	\caption{Three different attention insertion strategies in the DSE block are illustrated, corresponding to (a), (b), and (c), and their performance metrics are quantitatively compared in Table~\ref{tab:7}.}
	\label{Fig.8}
\end{figure}

To evaluate the impact of different attention insertion positions in the DSE block on model performance, we designed three insertion strategies: (a) Position1 (inserting the attention mechanism before channel merging), (b) Position2 (inserting the attention mechanism after channel merging), and (c) Position3 (inserting the attention mechanism separately into each branch), as illustrated in \text{Fig.~\ref{Fig.8}}. Table~\ref{tab:7} presents the ablation outcomes. The third strategy (c) demonstrated superior outcomes in every evaluated metric, recording a IoU of 68.05, a Pd of 95.24, and the minimal false alarm rate of 13.21, substantially surpassing the other two strategies. In contrast, although (a) and (b) achieved similar IoU values, they showed varying degrees of degradation in Pd and an increase in Fa, indicating that the insertion position of the attention mechanism has a significant impact on the model's feature representation ability. This demonstrates that inserting the attention at (c) helps preserve semantic integrity while effectively increasing the discriminability between targets and interference, which in turn improves detection performance.
\begin{table}[t]
	\renewcommand{\arraystretch}{1.2}
	\centering
	\setlength{\tabcolsep}{4pt} 
	\caption{Ablation study on attention insertion positions in the DSE block as measured through Params (M), FLOPs (G), \textit{IoU}~(\%), \textit{Pd}~(\%), and \textit{Fa}~($\times 10^{-6}$).}
	\label{tab:7}
	\resizebox{0.48\textwidth}{!}{%
		\begin{tabular}{c|c|c|c|c|c}
			\toprule
			\textbf{Position} & \textbf{Params (M)} & \textbf{FLOPs (G)} & \textbf{IoU$\uparrow$} & \textbf{P$_d\uparrow$} & \textbf{F$_a\downarrow$} \\
			\midrule
			(a) & \textbf{\textcolor{red}{4.4446}} & \textbf{\textcolor{red}{7.29465}} & 67.25 & 93.88 & 13.97 \\
			\rowcolor{gray!10}
			(b) & 4.4554 & 7.29468 & 67.96 & 93.54 & 14.50 \\
			\rowcolor{RoyalBlue!10}
			\textbf{(c)} & 4.4609 & 7.29665 & \textbf{\textcolor{red}{68.05}} & \textbf{\textcolor{red}{95.24}} & \textbf{\textcolor{red}{13.21}} \\
			\bottomrule
		\end{tabular}%
	} 
\end{table}
\subsubsection{Ablation Study on the Structural Design of the LaSEA Module}
To comprehensively assess the contribution of each component within the LaSEA module, we conducted component-level ablation tests. The IRSTD-1K benchmark results are documented in Table~\ref{tab:8}. First, removing the channel shuffle operation led to performance degradation: IoU dropped to 68.31, Pd to 91.84, and Fa increased to 9.57, indicating that channel shuffle helps enhance inter-channel interaction and reduce redundancy. Next, we independently evaluated the roles of the LaSEA module. Retaining only the cross-scale extraction resulted in a Fa value of 13.21, indicating that although multi-receptive field features are valuable, the lack of effective processing leads to a large amount of redundant features, which affects target saliency modeling. When the cross-scale path was removed, all three metrics showed a decrease to varying degrees, indicating that cross-scale semantics are crucial for enriching feature representation. We further combined the cross-scale extraction with other attention mechanisms, such as MaSA~\cite{fan2024rmt}, ESSA~\cite{zhang2023essaformer}, LSKe~\cite{li2023large}, and CRA~\cite{kang2024metaseg}, but these designs did not effectively handle deeper layers of the network. The findings validate the advantages of LaSEA design in enhancing deep target semantic regions and providing guidance for decoder fusion.
\begin{figure*}[]
	\centering
	\includegraphics[width=\textwidth]{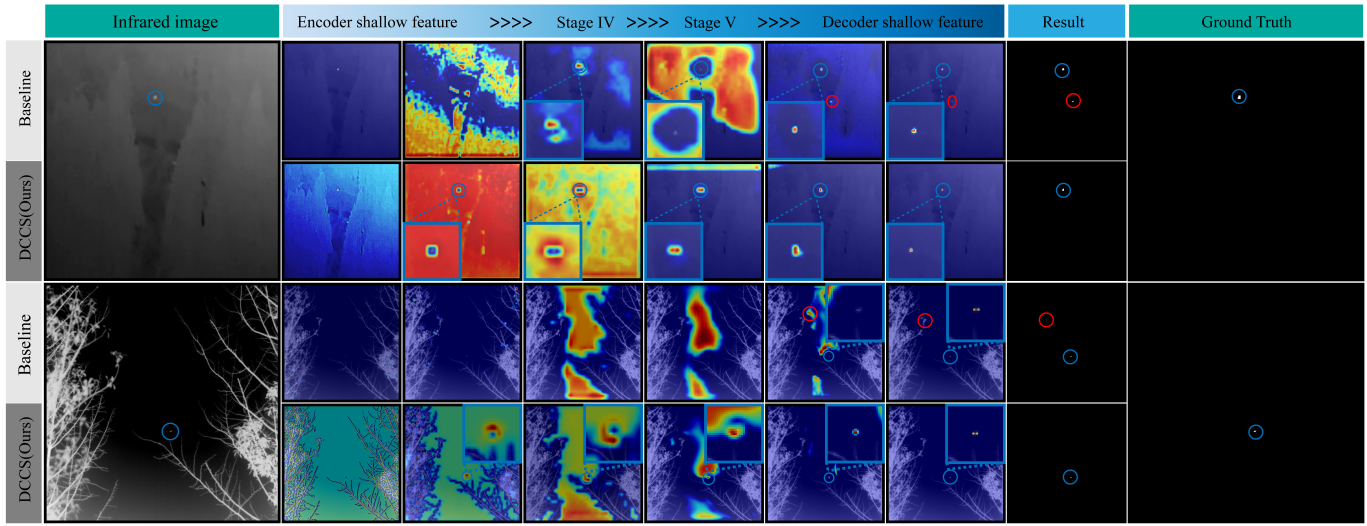}
	\caption{The figure presents a multi-stage visual comparison of features extracted by DCCS-Det and the baseline throughout the IRSTD pipeline, from the input image to the final output. Specifically, the encoder shallow features to Stage IV correspond to Stage I through IV in \text{Fig.~\ref{Fig.1}}, and the Stage V to decoder shallow features correspond to Stage V up to just before PD in \text{Fig.~\ref{Fig.1}}.}
	\label{Fig.9}
\end{figure*}
\begin{table}[t]
	\renewcommand{\arraystretch}{1.2}
	\centering
	\footnotesize 
	\setlength{\tabcolsep}{8.5pt} 
	\caption{Component-wise Ablation Study of the LaSEA Module on IRSTD-1K, showing \textit{IoU}~(\%), \textit{Pd}~(\%), and \textit{Fa}~($\times 10^{-6}$).}
	\label{tab:8}
	\resizebox{0.48\textwidth}{!}{%
		\begin{tabular}{lc|c|c|c}
			\toprule
			\textbf{Configuration} && \textbf{IoU$\uparrow$} & \textbf{P$_d\uparrow$} & \textbf{F$_a\downarrow$} \\
			\midrule
			Without Channel Shuffle && 68.31 & 91.84 & 9.57 \\
			\rowcolor{gray!10}
			Only Cross-scale && 67.25 & 92.18 & 13.21 \\
			Without Cross-scale && 68.28 & 91.84 & 9.57 \\
			\rowcolor{pink!5!white}  
			Cross-scale + MaSA~\cite{fan2024rmt}  && 67.73 & 91.16 & 10.86 \\
			Cross-scale + ESSA~\cite{zhang2023essaformer} && 66.76 & 91.16 & 16.40 \\
			\rowcolor{gray!10}
			Cross-scale + LSKe~\cite{li2023large} && 67.04 & 93.54 & 15.64 \\
			Cross-scale + CRA~\cite{kang2024metaseg} && 67.22 & 93.20 & 12.91 \\
			\rowcolor{RoyalBlue!10}
			\textbf{LaSEA(Ours)} && \textbf{\textcolor{red}{68.72}} & \textbf{\textcolor{red}{94.22}} & \textbf{\textcolor{red}{8.50}} \\		
			\bottomrule
		\end{tabular}
	}
\end{table}
\subsubsection{Ablation Study on the Pooling Strategy of the LaSEA Module}
To assess the efficacy of the random pooling strategy in the LaSEA module, we benchmark it against four deterministic pooling variants on the IRSTD-1K dataset, as reported in Table~\ref{tab:9}. Fixed-resolution pooling strategies (1×1, 2×2, and 3×3) each have their limitations: during training, they consistently allow the network to learn only scale-specific features, restricting the diversity of spatial information aggregation and thereby affecting effective guidance for decoder fusion. The ensemble strategy that averages across three resolutions forcibly fuses responses from different scales, where the fixed weighted fusion blurs the discriminative differences between scales, instead leading to a decrease in detection rate. In contrast, our random selection strategy achieves the best performance across all metrics. During training, the network dynamically selects randomly among different pooling scales, thereby preserving discriminative target features while serving as a Dropout-like regularization effect that encourages robust feature learning. During inference, deterministic 1×1 pooling is adopted to ensure stability while maintaining the generalization advantages learned during training.

\begin{table}[t]
	\renewcommand{\arraystretch}{1.2}
	\centering
	\footnotesize
	\setlength{\tabcolsep}{8.5pt}
	\caption{Comparison of Pooling Strategies in the LaSEA Module on IRSTD-1K, showing \textit{IoU}~(\%), \textit{Pd}~(\%), and \textit{Fa}~($\times 10^{-6}$).}
	\label{tab:9}
	\resizebox{0.48\textwidth}{!}{%
		\begin{tabular}{lc|c|c|c}
			\toprule
			\textbf{Pooling Strategy} && \textbf{IoU$\uparrow$} & \textbf{P$_d\uparrow$} & \textbf{F$_a\downarrow$} \\
			\midrule
			Fixed 1×1 (Global Pooling) && 67.50 & 93.88 & 12.07 \\
			\rowcolor{gray!10}
			Fixed 2×2 Pooling && 66.95 & 93.20 & 14.20 \\
			Fixed 3×3 Pooling && 67.48 & 92.86 & 10.93 \\
			\rowcolor{gray!10}
			Average of All Three && 67.44 & 91.84 & 11.31 \\
			\rowcolor{RoyalBlue!10}
			\textbf{Random Pooling (Ours)} && \textbf{\textcolor{red}{68.72}} & \textbf{\textcolor{red}{94.22}} & \textbf{\textcolor{red}{8.50}} \\		
			\bottomrule
		\end{tabular}
	}
\end{table}

\subsection{Analysis of multi-stage feature heatmaps}
To verify the effectiveness of the design, Fig.~\ref{Fig.9} presents a heatmap comparison between DCCS-Det and a conventional baseline method on the task of IRSTD. From left to right, each column corresponds to the input infrared image, shallow encoder features, features from Stage IV and Stage V, shallow decoder features, and the final detection result. As shown in the visualizations, DCCS-Det effectively captures global contextual dependencies and local details from shallow encoder features to Stage IV by jointly modeling the DSE block as an auxiliary branch and the primary branch, which allows for a clear semantic distinction between targets and interference. Subsequently, the LaSEA module further enhances the target semantic representation between Stage IV and Stage V, providing better guidance for the decoder and leading to improved detection accuracy. In contrast, the baseline method fails to adequately model contextual information, causing small targets to be easily overwhelmed by complex backgrounds. In addition, it does not mine deeper layers of the network, resulting in the inability to provide effective guidance for the decoder. The final detection results show that DCCS-Det achieves more stable and accurate small target detection under complex backgrounds, whereas the performance of the baseline method is relatively poor.
\section{CONCLUSION AND DISCUSSION}
This paper presents DCCS-Det for IRSTD. It employs a DSE block to construct an auxiliary branch, which supplements the primary branch by capturing local details and global context information, thereby reinforcing the boundary difference between the target and interference. We also design LaSEA module, which adopts cross-scale feature extraction and random pooling sampling strategies to alleviate deep network feature degradation, suppress noise, and enhance target semantic representation, thus guiding the decoder's fusion, leading to more accurate target detection. We conduct comprehensive evaluations using our approach across three public IRSTD datasets, demonstrating state-of-the-art detection accuracy with linear computational complexity, substantially outperforming Transformer-based hybrids in efficiency while remaining competitive with lightweight CNN baselines.

Despite achieving favorable trade-offs between accuracy and efficiency, our approach has certain limitations that may impede real-time deployment on resource-constrained platforms. Moreover, the current framework primarily focuses on spatial-domain feature extraction, without fully exploiting complementary information from other domains (e.g., frequency domain).

In future work, we plan to investigate lightweight network architectures for efficient edge deployment and explore multi-domain feature fusion strategies to further enhance the robustness and generalization capability of our method.
\bibliographystyle{IEEEtran}
\bibliography{reference} 
\end{document}